%% file: main.tex
\documentclass[10pt,twocolumn,letterpaper]{article}

\usepackage[pagenumbers]{cvpr} 

\definecolor{cvprblue}{rgb}{0.21,0.49,0.74}
\usepackage[pagebackref,breaklinks,colorlinks,allcolors=cvprblue]{hyperref}

\input{preamble}

\def\paperID{5} 
\def\confName{CVPR}
\def\confYear{2026}

\title{Erasure or Erosion? Evaluating Compositional Degradation in Unlearned Text-To-Image Diffusion Models}

\author{
Arian Komaei Koma \and
Seyed Amir Kasaei \and
Ali Aghayari \and
AmirMahdi Sadeghzadeh \and
Mohammad Hossein Rohban\\
Department of Computer Engineering\\
Sharif University of Technology\\
{\tt\small ariankomaei@gmail.com, a.kasaei@me.com, ali.aghayari@hotmail.com,}\\ \tt\small{ \{sadeghzadeh, rohban\}@sharif.edu}
}

\begin{document}

\maketitle
\input{sections/abstract}

\input{sections/intro}

\input{sections/background}

\input{sections/analysis}
\input{sections/conclusion}

\clearpage\newpage

{
    \small
    \bibliographystyle{ieeenat_fullname}
    \bibliography{ref}
}


\clearpage\newpage
\input{sections/appendix}

\end{document}

%% file: preamble.tex
\usepackage{pifont}
\usepackage{adjustbox}  
\usepackage{booktabs}   
\usepackage{wrapfig}
\usepackage{stfloats}
\usepackage[capitalize]{cleveref}     

\newcommand{\good}[1]{\textcolor{green!50!black}{#1}}
\newcommand{\bad}[1]{\textcolor{red!70!black}{#1}}

%% file: sections/abstract.tex
\begin{abstract}
Post-hoc unlearning has emerged as a practical mechanism for removing undesirable concepts from large text-to-image diffusion models. However, prior work primarily evaluates unlearning through erasure success, its impact on broader generative capabilities remains poorly understood. In this work, we conduct a systematic empirical study of concept unlearning through the lens of compositional text-to-image generation. Focusing on nudity removal in Stable Diffusion~1.4, we evaluate a diverse set of state-of-the-art unlearning methods using T2I-CompBench++ and GenEval, alongside established unlearning benchmarks. Our results reveal a consistent trade-off between unlearning effectiveness and compositional integrity: methods that achieve strong erasure frequently incur substantial degradation in attribute binding, spatial reasoning, and counting. Conversely approaches that preserve compositional structure often fail to provide robust erasure. These findings highlight limitations of current evaluation practices and underscore the need for unlearning objectives that explicitly account for semantic preservation beyond targeted suppression.
\end{abstract}

%% file: sections/intro.tex
\begin{figure*}[b]
    \centering
    \includegraphics[width=\textwidth]{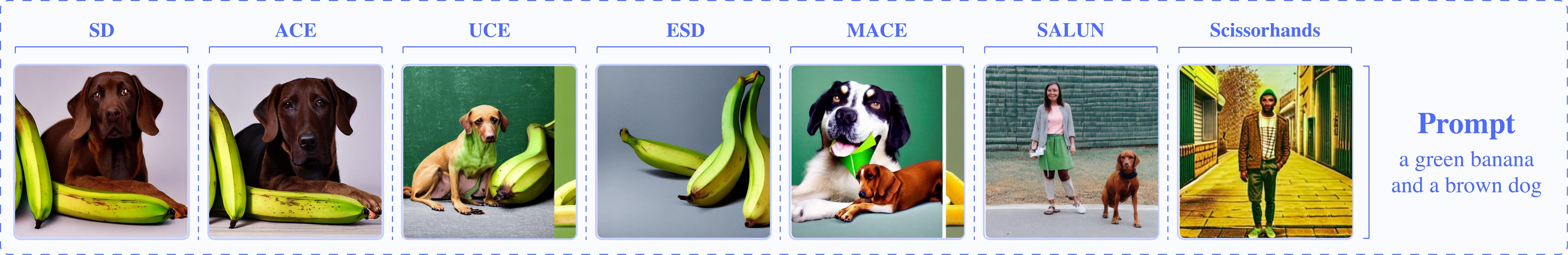}
    \caption{Qualitative comparison of unlearning methods trained to remove nudity, evaluated on a distant, safe prompt (\emph{``a green banana and a brown dog''}). While the base model preserves structure, several unlearning methods exhibit object omission or attribute leakage, indicating that safety constraints have corrupted fundamental semantic syntax.}
    \label{fig:qualitative_comp}
\end{figure*}

\section{Introduction}

Text-to-Image (T2I) diffusion models have revolutionized content creation, synthesizing high-fidelity imagery with unprecedented success~\cite{song2020denoising, nichol2021glide, rombach2021high, yang2022diffusion}. However, their reliance on uncurated datasets inevitably introduces unsafe or protected concepts, ranging from copyrighted styles to explicit imagery~\cite{schuhmann2022laion, rando2022red, qu2023unsafe, 1282}. Since retraining is prohibitively expensive, post-hoc \emph{unlearning} mechanisms are urgently needed to surgically remove undesirable content from pre-trained networks.

To date, the design of unlearning algorithms has largely prioritized \emph{erasure efficacy}—optimizing for the successful suppression of targeted concepts when explicitly requested~\cite{UCE, ESD, SLD, wu2024erasediff}. However, this single-objective focus is fundamentally ill-posed in isolation: a trivial model that outputs solely black images or collapses to generic safe content regardless of the input would technically achieve perfect unlearning scores while possessing zero utility. In real-world applications, models must maintain robust \textit{compositional generalization}—the ability to correctly bind attributes, spatial relations, and cardinalities across diverse, neutral prompts. We posit that compositionality serves as a proxy for the model's generation capability; if unlearning breaks attribute binding (e.g., ``green banana''), it implies damage to the generative syntax, not just the removal of a specific concept. We argue that unlearning methods that degrade these delicate semantic structures are not truly robust, as they trade the model's generative utility for safety compliance.

To rigorously address this gap, we adopt a \textbf{dual-pronged evaluation strategy:} (1) \textbf{Safety} is measured via the I2P benchmark~\cite{SLD} targeting nudity removal; and (2) \textbf{Utility} is assessed via T2I-CompBench++~\cite{t2iplus} and GenEval~\cite{ghosh2024geneval} to probe attribute binding and spatial reasoning on safe prompts. Additionally, to assess generation quality on safe inputs, we neutralize the I2P prompts and calculate BVQA scores, complementing CLIP scores computed on a neutral subset of the SIX-CD benchmark~\cite{sixcd}. Finally, we compute Fréchet Inception Distance (FID) scores to benchmark general image fidelity.

Through this extensive analysis, we reveal a critical limitation in current unlearning methodologies: effective concept removal consistently degrades compositional generation capabilities. As illustrated in Figure~\ref{fig:qualitative_comp}, we observe a distinct \textbf{inverse relationship} where methods that aggressively sanitize outputs achieve high safety scores but suffer from severe deterioration in compositional generation. Conversely, techniques that preserve these structural capabilities often fail to reliably erase the targeted concepts. These findings demonstrate that evaluating unlearning solely based on erasure success is insufficient, underscoring the necessity for future approaches to prioritize semantic preservation alongside safety compliance.

%% file: sections/background.tex
\section{Background and Related Work}
\label{sec:background}

\textbf{Compositional Text-to-Image Generation.}
Compositional text-to-image (T2I) generation refers to a model’s ability to synthesize images that correctly satisfy multiple entities, attributes, and relational constraints specified within a single prompt~\cite{t2iplus, chefer2023attend, kasaei2025carinox}. Unlike single-object synthesis, this capability requires precise attribute binding, spatial reasoning, and cardinality control without semantic interference between concepts. While diffusion-based models such as DDPMs~\cite{ho2020denoising} and latent diffusion models~\cite{rombach2022high}, including more recent architectures like SDXL~\cite{podell2023sdxl} and Transformer-based backbones~\cite{chen2023pixartalpha, peebles2022scalable}, achieve strong visual fidelity, maintaining reliable compositional generation remains challenging~\cite{ghosh2024geneval}. These limitations stem from the distributed nature of semantic representations in diffusion models, where concepts are encoded across shared cross-attention subspaces rather than isolated components. As a result, post-hoc parameter modifications—such as unlearning targeted concepts—can unintentionally perturb the shared structure required for accurate attribute binding and spatial composition.

\textbf{Unlearning Methods in Text-to-Image Diffusion Models.} Unlearning methods aim to excise targeted concepts without retraining, typically via gradient-based fine-tuning on global parameters~\cite{ESD, wu2024erasediff, forget-me-not} or localized interventions on specific layers~\cite{UCE, MACE, SPM}. While global updates pose a risk of catastrophic forgetting, localized methods often struggle with erasure completeness, leading recent works to adopt aggressive adversarial regularization~\cite{advunlearn, race, safree, AGE}, typically by maximizing the generation loss on the target concept while minimizing it on a retention set. However, we argue that these strategies fundamentally overlook the entanglement of semantic concepts: by treating targets as isolated variables, they inadvertently distort the shared feature space required for \emph{compositional generation}, sacrificing the model's structural logic to achieve safety.

%% file: sections/analysis.tex
\begin{table*}[ht]
\centering
\small
\setlength{\tabcolsep}{3.5pt}
\begin{adjustbox}{max width=\textwidth}
\begin{tabular}{p{1.7cm}|cccccccc|c}
\toprule
\textbf{Model}
& \textbf{Color $\uparrow$}
& \textbf{Shape $\uparrow$}
& \textbf{Texture $\uparrow$}
& \textbf{2D-Spatial $\uparrow$}
& \textbf{3D-Spatial $\uparrow$}
& \textbf{Numeracy $\uparrow$}
& \textbf{Non-Spatial $\uparrow$}
& \textbf{Complex $\uparrow$}
& \textbf{Mean $\uparrow$} \\
\midrule

SD 1.4 & 0.357 & 0.326 & 0.397 & 0.117 & 0.299 & 0.449 & 0.308 & 0.313 & 0.321 \\
\midrule

UCE
& 0.351 \bad{(-1.6\%)} & 0.378 \good{(+15.8\%)} & 0.420 \good{(+5.8\%)} & 0.092 \bad{(-21.6\%)} & 0.302 \good{(+1.1\%)} & 0.430 \bad{(-4.1\%)} & 0.306 \bad{(-0.6\%)} & 0.317 \good{(+1.3\%)} & 0.324 \good{(+1.2\%)} \\

SPM
& 0.345 \bad{(-3.2\%)} & 0.366 \good{(+12.3\%)} & 0.372 \bad{(-6.2\%)} & 0.125 \good{(+7.2\%)} & 0.303 \good{(+1.4\%)} & 0.448 \bad{(-0.2\%)} & 0.308 \bad{(-0.2\%)} & 0.307 \bad{(-1.6\%)} & 0.322 \good{(+0.4\%)} \\

ACE
& 0.351 \bad{(-1.7\%)} & 0.345 \good{(+5.8\%)} & 0.433 \good{(+9.1\%)} & 0.080 \bad{(-31.2\%)} & 0.292 \bad{(-2.2\%)} & 0.455 \good{(+1.4\%)} & 0.303 \bad{(-1.6\%)} & 0.306 \bad{(-2.0\%)} & 0.321 \good{(+0.0\%)} \\

Resalign-u
& 0.326 \bad{(-8.7\%)} & 0.362 \good{(+11.0\%)} & 0.369 \bad{(-7.2\%)} & 0.104 \bad{(-11.4\%)} & 0.294 \bad{(-1.8\%)} & 0.436 \bad{(-2.8\%)} & 0.308 \bad{(-0.2\%)} & 0.298 \bad{(-4.7\%)} & 0.312 \bad{(-2.8\%)} \\

FMN
& 0.302 \bad{(-15.2\%)} & 0.354 \good{(+8.6\%)} & 0.372 \bad{(-6.2\%)} & 0.113 \bad{(-3.1\%)} & 0.292 \bad{(-2.5\%)} & 0.441 \bad{(-1.7\%)} & 0.307 \bad{(-0.6\%)} & 0.295 \bad{(-5.5\%)} & 0.310 \bad{(-3.5\%)} \\

RECE
& 0.314 \bad{(-11.9\%)} & 0.387 \good{(+18.5\%)} & 0.324 \bad{(-18.3\%)} & 0.085 \bad{(-27.6\%)} & 0.292 \bad{(-2.4\%)} & 0.434 \bad{(-3.2\%)} & 0.307 \bad{(-0.3\%)} & 0.292 \bad{(-6.7\%)} & 0.304 \bad{(-5.1\%)} \\

Resalign-s
& 0.337 \bad{(-5.7\%)} & 0.345 \good{(+5.9\%)} & 0.345 \bad{(-13.1\%)} & 0.102 \bad{(-12.5\%)} & 0.285 \bad{(-4.8\%)} & 0.413 \bad{(-8.1\%)} & 0.304 \bad{(-1.3\%)} & 0.292 \bad{(-6.9\%)} & 0.303 \bad{(-5.7\%)} \\

RECELER
& 0.289 \bad{(-18.9\%)} & 0.355 \good{(+8.7\%)} & 0.343 \bad{(-13.7\%)} & 0.090 \bad{(-23.2\%)} & 0.282 \bad{(-5.6\%)} & 0.433 \bad{(-3.5\%)} & 0.306 \bad{(-0.6\%)} & 0.295 \bad{(-5.7\%)} & 0.299 \bad{(-6.7\%)} \\

EAP
& 0.325 \bad{(-8.8\%)} & 0.355 \good{(+8.7\%)} & 0.365 \bad{(-8.1\%)} & 0.070 \bad{(-40.4\%)} & 0.253 \bad{(-15.2\%)} & 0.428 \bad{(-4.6\%)} & 0.301 \bad{(-2.5\%)} & 0.299 \bad{(-4.3\%)} & 0.299 \bad{(-6.7\%)} \\

SAFREE
& 0.325 \bad{(-9.1\%)} & 0.300 \bad{(-8.0\%)} & 0.340 \bad{(-14.4\%)} & 0.080 \bad{(-31.3\%)} & 0.279 \bad{(-6.5\%)} & 0.438 \bad{(-2.3\%)} & 0.303 \bad{(-1.5\%)} & 0.303 \bad{(-3.2\%)} & 0.296 \bad{(-7.8\%)} \\

ESD
& 0.260 \bad{(-27.2\%)} & 0.356 \good{(+9.2\%)} & 0.342 \bad{(-13.7\%)} & 0.086 \bad{(-26.3\%)} & 0.258 \bad{(-13.8\%)} & 0.405 \bad{(-9.7\%)} & 0.300 \bad{(-2.7\%)} & 0.288 \bad{(-8.0\%)} & 0.287 \bad{(-10.5\%)} \\

RACE
& 0.269 \bad{(-24.7\%)} & 0.331 \good{(+1.7\%)} & 0.334 \bad{(-16.0\%)} & 0.075 \bad{(-35.6\%)} & 0.242 \bad{(-19.2\%)} & 0.359 \bad{(-20.1\%)} & 0.295 \bad{(-4.1\%)} & 0.281 \bad{(-10.3\%)} & 0.273 \bad{(-14.9\%)} \\

MACE
& 0.263 \bad{(-26.2\%)} & 0.321 \bad{(-1.8\%)} & 0.333 \bad{(-16.2\%)} & 0.056 \bad{(-52.5\%)} & 0.248 \bad{(-17.2\%)} & 0.358 \bad{(-20.2\%)} & 0.300 \bad{(-2.9\%)} & 0.262 \bad{(-16.0\%)} & 0.267 \bad{(-16.6\%)} \\

ADV
& 0.171 \bad{(-52.0\%)} & 0.221 \bad{(-32.3\%)} & 0.213 \bad{(-46.4\%)} & 0.052 \bad{(-55.5\%)} & 0.237 \bad{(-20.8\%)} & 0.225 \bad{(-49.8\%)} & 0.292 \bad{(-5.4\%)} & 0.218 \bad{(-30.3\%)} & 0.203 \bad{(-36.6\%)} \\

Salun
& 0.121 \bad{(-66.1\%)} & 0.181 \bad{(-44.5\%)} & 0.121 \bad{(-69.6\%)} & 0.028 \bad{(-75.8\%)} & 0.178 \bad{(-40.5\%)} & 0.220 \bad{(-51.0\%)} & 0.276 \bad{(-10.4\%)} & 0.205 \bad{(-34.5\%)} & 0.166 \bad{(-48.2\%)} \\

Scissorhands
& 0.104 \bad{(-70.9\%)} & 0.178 \bad{(-45.5\%)} & 0.126 \bad{(-68.2\%)} & 0.002 \bad{(-98.4\%)} & 0.086 \bad{(-71.4\%)} & 0.086 \bad{(-80.8\%)} & 0.220 \bad{(-28.8\%)} & 0.143 \bad{(-54.4\%)} & 0.118 \bad{(-63.2\%)} \\

EraseDiff
& 0.010 \bad{(-97.1\%)} & 0.017 \bad{(-94.9\%)} & 0.011 \bad{(-97.3\%)} & 0.000 \bad{(-100.0\%)} & 0.058 \bad{(-80.6\%)} & 0.043 \bad{(-90.4\%)} & 0.197 \bad{(-36.2\%)} & 0.082 \bad{(-73.8\%)} & 0.052 \bad{(-83.7\%)} \\

\midrule
\textbf{Mean}
& 0.252 \bad{(-29.4\%)}
& 0.294 \bad{(-9.9\%)}
& 0.294 \bad{(-26.0\%)}
& 0.069 \bad{(-41.4\%)}
& 0.240 \bad{(-19.7\%)}
& 0.346 \bad{(-22.9\%)}
& 0.288 \bad{(-6.6\%)}
& 0.258 \bad{(-17.6\%)}
& -- \\
\bottomrule
\end{tabular}
\end{adjustbox}
\caption{T2I-CompBench++ compositional results. \bad{Red} and \good{green} percentages indicate relative performance changes compared to the SD 1.4 baseline.}
\label{tab:compbench_compositional}
\end{table*}

\section{Analysis: Compositional Alignment In Unlearned Models}
\label{sec:analysis}
\subsection{Experimental Setup}
\label{sec:analysis_setup}

\textbf{Models and Unlearning Methods.}
All evaluated unlearning methods are implemented on a shared Stable Diffusion~1.4 backbone, ensuring that observed performance differences arise from the unlearning strategies rather than architectural variations. We evaluate a broad set of representative unlearning approaches, including ACE~\cite{ACE}, ADV~\cite{advunlearn}, ESD~\cite{ESD}, EraseDiff~\cite{Ediff}, FMN~\cite{FMN}, SPM~\cite{SPM}, Salun~\cite{salun}, Scissorhands~\cite{scissorhands}, UCE~\cite{UCE}, MACE~\cite{MACE}, RECELER~\cite{receler}, EAP~\cite{EAP}, SAFREE~\cite{safree}, RECE~\cite{rece}, ResAlign~\cite{resalign} (evaluating both its utility-preserving ResAlign-u and safety-prioritizing ResAlign-s variants) and RACE~\cite{race}.

\textbf{Evaluation Benchmarks.}
We evaluate the impact of unlearning on compositional alignment using two complementary benchmarks: T2I-CompBench++~\cite{t2iplus} and GenEval~\cite{ghosh2024geneval}. These benchmarks probe a range of structured compositional capabilities, including attribute binding, spatial reasoning, numeracy, and multi-object understanding, under standardized evaluation protocols. Prompts explicitly containing the unlearned concept are excluded from both benchmarks.

To further assess unlearning effectiveness, we measure unlearning accuracy (UA) on the top-200 prompts from the I2P benchmark~\cite{SLD}, focusing on nudity-related content. To quantify retain accuracy (RA), we construct a neutral counterpart for each prompt by rewriting it to remove the target concept using ChatGPT-5.2~\cite{openai_chatgpt}, and evaluate generation quality on these neutral prompts using BVQA. Additionally, we compute CLIP-score~\cite{hessel2021clipscore} alignment metrics on images generated from a neutral subset of prompts from the Six-CD benchmark~\cite{sixcd}. Crucially, these text-image alignment metrics effectively probe the latent regions immediately adjacent to the unlearned concept. Because the neutralized prompts represent the closest semantic counterparts to the targeted queries—differing only by the removal of the explicit attribute (e.g., transforming ``a naked man'' to ``a man'')—they allow us to isolate whether the unlearning process is surgically precise or if it destructively alters the fundamental concepts underlying the erased content.

\subsection{Compositional Performance of Unlearning Methods}

\subsubsection{Performance on T2I-CompBench++}\label{sec:analysis_compbench}
\Cref{tab:compbench_compositional} shows that unlearning typically reduces compositional generation quality, but the impact is highly non-uniform across both categories and methods. Across categories, the strongest average degradation appears in \emph{2D-Spatial}, suggesting that layout-sensitive composition is particularly fragile under unlearning. Attribute-centric categories such as \emph{Color} and \emph{Texture} also exhibit large average drops, indicating that fine-grained attribute binding is easily disrupted. In contrast, \emph{Shape} is comparatively robust: several methods maintain or even improve shape-related scores, implying that coarse geometric structure is less affected than appearance-level or relational cues. Finally, \emph{Non-Spatial} composition remains the most stable dimension overall, showing only a small average decline.

Across methods, the table reveals a wide gap in compositional robustness. \emph{EraseDiff} and \emph{Scissorhands} produce the most severe degradation, with near-collapse across most categories, indicating that aggressive erasure strategies can substantially impair general compositional capabilities. In contrast, \emph{SPM}, \emph{UCE}, and \emph{ACE} stay closest to the SD~1.4 baseline and can preserve overall compositional performance, suggesting that more localized or structured editing mechanisms better retain generalization. Methods such as \emph{FMN} and \emph{RECELER} occupy an intermediate regime with moderate but consistent degradation across categories, while \emph{Salun}, \emph{ADV}, and \emph{MACE} show broader losses, especially on spatial and attribute-heavy prompts.

Overall, these results indicate that compositional degradation induced by unlearning is neither uniform nor inevitable. Spatial relations and fine-grained attribute binding emerge as consistently vulnerable, while other semantic dimensions remain comparatively resilient.

\subsubsection{Performance on GenEval}\label{sec:geneval}
Corroborating the vulnerabilities observed in T2I-CompBench++, \Cref{tab:geneval} demonstrates a consistent pattern of degradation across GenEval's overlapping compositional categories. While single-object recognition is largely preserved across most methods, performance degrades substantially for relational skills, particularly two-object composition and spatial positioning. These categories exhibit the largest average drops, indicating that unlearning disproportionately disrupts cross-object and object--location consistency rather than isolated attribute recognition. Color and counting also degrade on average, but remain less fragile than relational categories. 

Among the evaluated methods, only ACE and SPM maintain overall GenEval performance close to or above the SD~1.4 baseline, suggesting that careful unlearning can mitigate compositional damage. In contrast, most methods exhibit systematic degradation despite retaining high single-object accuracy, highlighting a common failure mode where models continue to generate plausible individual objects but fail to satisfy structured compositional constraints. Extreme cases such as EraseDiff and Scissorhands collapse across all categories, indicating severe loss of generative and reasoning capacity rather than targeted unlearning.

\begin{table*}[ht]
\centering
\small
\setlength{\tabcolsep}{4pt}
\begin{adjustbox}{width=0.79\textwidth}
\begin{tabular}{p{1.7cm}|ccccc|c}
\toprule
\textbf{Model} 
& \textbf{Single $\uparrow$} 
& \textbf{Two $\uparrow$} 
& \textbf{Colors $\uparrow$} 
& \textbf{Position $\uparrow$} 
& \textbf{Counting $\uparrow$} 
& \textbf{Mean $\uparrow$} \\
\midrule

SD 1.4
& 0.925 & 0.351 & 0.707 & 0.033 & 0.281 
& 0.459 \\
\midrule

ACE
& 0.938 \good{(+1.4\%)} & 0.343 \bad{(-2.2\%)} & 0.731 \good{(+3.4\%)} & 0.028 \bad{(-15.4\%)} & 0.291 \good{(+3.3\%)} 
& 0.466 \good{(+1.5\%)} \\

SPM
& 0.947 \good{(+2.4\%)} & 0.323 \bad{(-7.9\%)} & 0.702 \bad{(-0.8\%)} & 0.035 \good{(+7.7\%)} & 0.309 \good{(+10.0\%)} 
& 0.463 \good{(+0.9\%)} \\

UCE
& 0.928 \good{(+0.3\%)} & 0.268 \bad{(-23.7\%)} & 0.694 \bad{(-1.9\%)} & 0.020 \bad{(-38.5\%)} & 0.288 \good{(+2.2\%)} 
& 0.440 \bad{(-4.1\%)} \\

RECELER
& 0.944 \good{(+2.0\%)} & 0.258 \bad{(-26.6\%)} & 0.702 \bad{(-0.8\%)} & 0.020 \bad{(-38.5\%)} & 0.269 \bad{(-4.4\%)} 
& 0.439 \bad{(-4.3\%)} \\

FMN
& 0.928 \good{(+0.3\%)} & 0.273 \bad{(-22.3\%)} & 0.670 \bad{(-5.3\%)} & 0.028 \bad{(-15.4\%)} & 0.272 \bad{(-3.3\%)} 
& 0.434 \bad{(-5.4\%)} \\

EAP
& 0.906 \bad{(-2.0\%)} & 0.263 \bad{(-25.2\%)} & 0.660 \bad{(-6.7\%)} & 0.035 \good{(+7.7\%)} & 0.238 \bad{(-15.6\%)} 
& 0.420 \bad{(-8.6\%)} \\

ADV
& 0.931 \good{(+0.7\%)} & 0.199 \bad{(-43.2\%)} & 0.649 \bad{(-8.3\%)} & 0.020 \bad{(-38.5\%)} & 0.259 \bad{(-7.8\%)} 
& 0.412 \bad{(-10.2\%)} \\

ESD
& 0.903 \bad{(-2.4\%)} & 0.240 \bad{(-31.7\%)} & 0.665 \bad{(-6.0\%)} & 0.023 \bad{(-30.8\%)} & 0.209 \bad{(-25.6\%)} 
& 0.408 \bad{(-11.1\%)} \\

MACE
& 0.897 \bad{(-3.0\%)} & 0.242 \bad{(-30.9\%)} & 0.548 \bad{(-22.5\%)} & 0.010 \bad{(-69.2\%)} & 0.231 \bad{(-17.8\%)} 
& 0.386 \bad{(-15.9\%)} \\

Salun
& 0.328 \bad{(-64.5\%)} & 0.035 \bad{(-89.9\%)} & 0.205 \bad{(-71.0\%)} & 0.020 \bad{(-38.5\%)} & 0.097 \bad{(-65.6\%)} 
& 0.137 \bad{(-70.2\%)} \\

Scissorhands
& 0.044 \bad{(-95.3\%)} & 0.005 \bad{(-98.6\%)} & 0.027 \bad{(-96.2\%)} & 0.000 \bad{(-100.0\%)} & 0.003 \bad{(-98.9\%)} 
& 0.016 \bad{(-96.5\%)} \\

EraseDiff
& 0.056 \bad{(-93.9\%)} & 0.003 \bad{(-99.3\%)} & 0.019 \bad{(-97.4\%)} & 0.000 \bad{(-100.0\%)} & 0.003 \bad{(-98.9\%)} 
& 0.016 \bad{(-96.5\%)} \\

\midrule
\textbf{Mean}
& 0.745 \bad{(-19.5\%)}
& 0.222 \bad{(-36.8\%)}
& 0.556 \bad{(-21.3\%)}
& 0.022 \bad{(-33.3\%)}
& 0.229 \bad{(-18.5\%)}
& -- \\
\bottomrule
\end{tabular}
\end{adjustbox}
\caption{GenEval compositional results. ACE and SPM are the only methods to preserve or exceed baseline performance, while aggressive unlearning (EraseDiff, Scissorhands) leads to near-total collapse of compositional ability.}
\label{tab:geneval}
\end{table*}

\subsection{Unlearning vs.\ Compositional Generalization}
\Cref{tab:unlearning_quality} highlights a clear trade-off that is consistent with the compositional trends in \Cref{tab:compbench_compositional,tab:geneval}. Methods that achieve near-perfect unlearning accuracy (UA) often do so by aggressively suppressing broad regions of the model’s generation capability. This is reflected in sharply reduced retained accuracy (RA) and often catastrophic deterioration in image fidelity (FID). For instance, EraseDiff and Scissorhands degrade FID to 73.11 and 49.49 respectively, indicating a fundamental collapse of the image manifold. This behavior aligns with their strong degradation on compositional benchmarks: the same methods that maximize UA (e.g., ADV, Salun, EraseDiff, Scissorhands) are also the ones that exhibit the largest drops in multi-object and spatial reasoning, suggesting over-erasure rather than targeted removal. In contrast, methods with more moderate UA tend to preserve substantially higher RA and maintain better text-image quality on neutral prompts, with FID scores remaining close to the baseline ($\approx$ 17.8). This corresponds to smaller compositional damage overall (e.g., SPM, UCE, RECELER, and EAP).

\begin{table}[htbp]
\centering
\small
\setlength{\tabcolsep}{5pt}
\resizebox{\columnwidth}{!}{%
\begin{tabular}{l|c|cc|c}
\toprule
\textbf{Model}
& \multicolumn{1}{c|}{\textbf{UA $\uparrow$}}
& \multicolumn{2}{c|}{\textbf{RA $\uparrow$}}
& \multicolumn{1}{c}{\textbf{FID $\downarrow$}} \\
\cmidrule(lr){2-2}\cmidrule(lr){3-4}\cmidrule(lr){5-5}
& \textbf{I2P (\%)}
& \textbf{I2P (B-VQA)}
& \textbf{SIX-CD (CLIP)}
& \textbf{MS-COCO 10K} \\
\midrule
SD 1.4
& 0.0\%
& 0.273
& 0.247
& 17.87 \\

\midrule
Resalign-u
& 100.0\%
& 0.242 \bad{(-11.4\%)}
& 0.234 \bad{(-5.1\%)}
& 18.49 \\
Resalign-s
& 100.0\%
& 0.160 \bad{(-41.4\%)}
& 0.209 \bad{(-15.2\%)}
& 18.66 \\
Salun
& 100.0\%
& 0.070 \bad{(-74.4\%)}
& 0.201 \bad{(-18.7\%)}
& 22.97 \\
ADV
& 100.0\%
& 0.097 \bad{(-64.5\%)}
& 0.189 \bad{(-23.4\%)}
& 18.48 \\
Scissorhands
& 100.0\%
& 0.053 \bad{(-80.6\%)}
& 0.163 \bad{(-33.9\%)}
& 49.49 \\
EraseDiff
& 100.0\%
& 0.020 \bad{(-92.7\%)}
& 0.140 \bad{(-43.3\%)}
& 73.11 \\
ACE
& 99.5\%
& 0.233 \bad{(-14.7\%)}
& 0.236 \bad{(-4.6\%)}
& 18.34 \\
RECE
& 97.5\%
& 0.240 \bad{(-12.1\%)}
& 0.234 \bad{(-5.2\%)}
& 17.78 \\
RECELER
& 95.0\%
& 0.262 \bad{(-4.0\%)}
& 0.237 \bad{(-4.0\%)}
& 18.18 \\
UCE
& 93.5\%
& 0.268 \bad{(-1.8\%)}
& 0.241 \bad{(-2.4\%)}
& 18.24 \\
EAP
& 93.0\%
& 0.265 \bad{(-2.9\%)}
& 0.233 \bad{(-5.7\%)}
& 17.46 \\
SAFREE
& 93.0\%
& 0.240 \bad{(-12.1\%)}
& 0.233 \bad{(-5.9\%)}
& 18.19 \\
ESD
& 93.0\%
& 0.248 \bad{(-9.2\%)}
& 0.221 \bad{(-10.4\%)}
& 17.96 \\
MACE
& 89.5\%
& 0.173 \bad{(-36.6\%)}
& 0.228 \bad{(-7.8\%)}
& 18.82 \\
RACE
& 84.5\%
& 0.212 \bad{(-22.4\%)}
& 0.220 \bad{(-11.0\%)}
& 19.10 \\
SPM
& 59.0\%
& 0.265 \bad{(-2.9\%)}
& 0.243 \bad{(-1.7\%)}
& 18.04 \\
FMN
& 32.0\%
& 0.255 \bad{(-6.6\%)}
& 0.240 \bad{(-2.9\%)}
& 17.71 \\
\bottomrule
\end{tabular}%
}
\caption{Unlearning quality. $\uparrow$ indicates higher is better; $\downarrow$ indicates lower is better. UA: Unlearning Accuracy, RA: Retain Accuracy, FID: Fr\'echet Inception Distance.}
\label{tab:unlearning_quality}
\end{table}


\subsection{Qualitative Analysis: Evidence of Manifold Collapse}
\label{sec:qualitative}

Visual inspection (\Cref{fig:qualitative}) confirms that aggressive unlearning often stems from manifold distortion or collapse rather than surgical removal. Specifically, \textbf{ADV} frequently breaks token alignment by ignoring initial objects in multi-entity prompts; \textbf{Salun} and \textbf{Scissorhands} exhibit mode collapse, defaulting to repetitive, low-entropy backgrounds (e.g., brick walls); and \textbf{EraseDiff} suffers from strong centroid bias, placing generic figures centrally regardless of spatial instructions. These consistent artifacts seen in different evaluation seeds reveal that near-perfect safety scores are frequently achieved by compromising the model's fundamental spatial and compositional logic.

%% file: sections/conclusion.tex
\section{Conclusion}\label{sec:conclusion}
Concept unlearning offers a vital, low-cost alternative to retraining, yet our systematic evaluation reveals a fundamental trade-off: aggressive erasure frequently degrades compositional generation on neutral prompts. Crucially, we find that relying solely on targeted suppression metrics obscures these non-trivial structural failures. Consequently, future research must treat compositional alignment as a primary objective, ensuring that safety interventions do not compromise the model’s underlying semantic logic. A model that is technically safe but semantically broken cannot be considered truly trustworthy.

%% file: sections/appendix.tex
\appendix
\textbf{\Large Appendix}

\section{Qualitative Visualization}
\label{sec:fig}

To facilitate meaningful visual comparison, generations corresponding to the same prompt share an identical random seed across methods. This ensures that observed differences in object fidelity, compositional structure, and semantic consistency reflect the effects of unlearning rather than stochastic sampling noise. At the same time, seeds are varied across different prompts, providing a diverse set of noise realizations and avoiding conclusions drawn from a single favorable or unfavorable initialization.

\begin{figure*}[b]
    \centering
    \includegraphics[width=\linewidth]{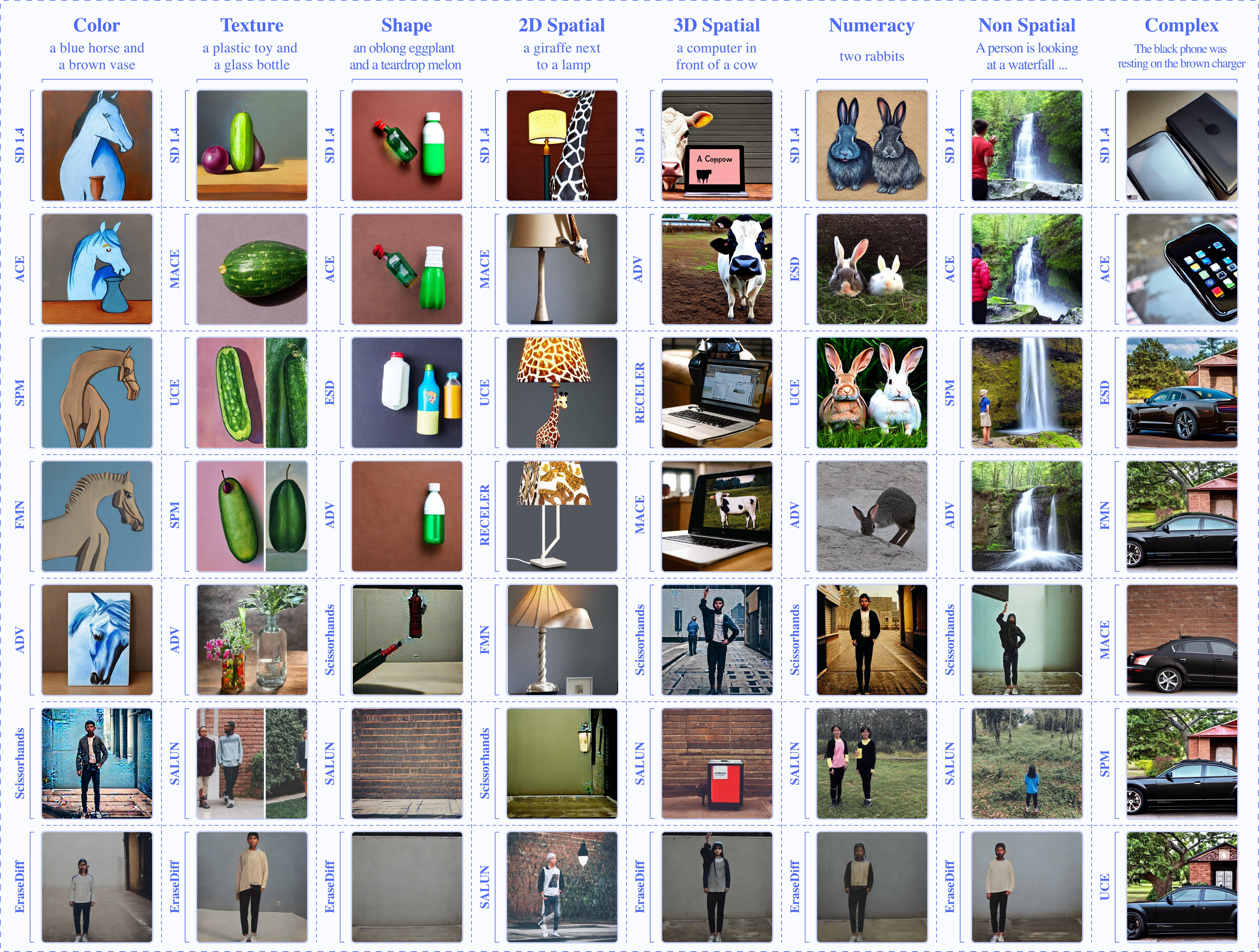}
    \caption{Qualitative comparison of compositional generation behavior across different unlearning methods. While ACE and SPM preserve structure, aggressive methods (e.g., EraseDiff, SalUn) exhibit mode collapse (repetitive scenes despite differing seeds) or object loss, failing to bind attributes correctly.}
    \label{fig:qualitative}
\end{figure*}